\documentclass{article}

\PassOptionsToPackage{round}{natbib}
\usepackage[preprint]{neurips_2024}

\usepackage[utf8]{inputenc} 
\usepackage[T1]{fontenc}    
\usepackage{hyperref}       
\usepackage{url}            
\usepackage{booktabs}       
\usepackage{amsfonts}       
\usepackage{nicefrac}       
\usepackage{microtype}      
\usepackage{xcolor}         

\usepackage{graphicx}
\usepackage{caption}
\usepackage{subcaption}
\usepackage{dsfont}
\usepackage{amsmath}
\usepackage{amssymb}
\usepackage{amsthm}
\usepackage{enumitem}
\usepackage{placeins}
\usepackage{verbatim}
\usepackage[ruled,vlined]{algorithm2e}

\DeclareMathOperator*{\argmin}{argmin}

\renewcommand{\d}[0]{\mathrm{d}}

\newcommand{\RR}{\mathbb{R}}

\newcommand{\smallp}[1]{{{\scriptscriptstyle(}#1{\scriptscriptstyle)}}}

\newcommand{\EE}{\mathbb{E}}

\newcommand{\ud}{\, {\rm d} \kern-.015em }

\newcommand{\modulus}[1]{\left| \kern.05em #1 \kern.05em \right|}
\newcommand{\norm}[1]{\left\| \kern.05em #1 \kern.05em \right\|}
\newcommand{\inner}[1]{\left\langle \kern.05em #1 \kern.05em \right\rangle }

\newcommand{\pick}[2]{\renewcommand{\arraystretch}{0.6}
	\left( \kern-.4em \begin{array}{c} #1 \\ #2 \end{array} \kern-.4em \right) }

\newcommand{\Var}{\, {\rm Var}}

\title{Diffusion Generative Modelling for Divide-and-Conquer MCMC}

\author{
  Connie ~Trojan\\
  Department of Mathematics and Statistics\\
  Lancaster University\\
  Lancaster, LA1 4YF, UK \\
  \texttt{c.trojan1@lancaster.ac.uk}\\
  \And 
  Paul ~Fearnhead\\
  Department of Mathematics and Statistics\\
  Lancaster University\\
  Lancaster, LA1 4YF, UK \\
  \texttt{p.fearnhead@lancaster.ac.uk}\\
  \And 
  Christopher ~Nemeth\\
  Department of Mathematics and Statistics\\
  Lancaster University\\
  Lancaster, LA1 4YF, UK \\
  \texttt{c.nemeth@lancaster.ac.uk}\\
}

\begin{document}

\maketitle

\begin{abstract}
Divide-and-conquer MCMC is a strategy for parallelising Markov Chain Monte Carlo sampling by running independent samplers on disjoint subsets of a dataset and merging their output. An ongoing challenge in the literature is to efficiently perform this merging without imposing distributional assumptions on the posteriors. We propose using diffusion generative modelling to fit density approximations to the subposterior distributions. This approach outperforms existing methods on challenging merging problems, while its computational cost scales more efficiently to high dimensional problems than existing density estimation approaches.
\end{abstract}


\section{Introduction}

Markov chain Monte Carlo samplers are a common numerical tool for Bayesian inference. However, each update step of a Metropolis-Hastings MCMC sampler requires a calculation involving the full dataset to compute the acceptance probability. This gives each iteration of the MCMC algorithm a linear complexity in the number of datapoints, which is computationally impractical for very large datasets, thus motivating the use of faster approximate samplers.

There are two general approaches to MCMC to reduce the per-iteration cost \citep[see][for an overview]{Bardenetetal2017tallMCMC}. One is based on approximate MCMC algorithms that use a subset of the data at each iteration. Such methods include stochastic gradient Langevin dynamics \citep{WellingTeh2011} or its extensions \citep{nemeth2021stochastic}. The other is a divide-and-conquer approach \citep{Scottetal2016consensus,Neiswangeretal2014parallel}, where the dataset is partitioned before inference so that MCMC chains can be run in parallel on each subset of the data and then merged. This latter approach has several advantages over the former, as we can trivially run MCMC algorithms on data subsets in an embarrassingly parallel fashion, across multiple CPUs, without incurring a communication cost between CPUs. However, these methods are less commonly used in practice due to the challenge of reliably merging the information from the MCMC output on different data subsets. Whilst our work is motivated by divide-and-conquer MCMC, similar challenges of merging information from disjoint datasets arise in federated learning \citep[see e.g.][]{Lietal2020FL}, where the data is naturally partitioned into subsets that cannot be merged due to privacy or communication constraints. The approach we present to solve the challenge of parallel MCMC could also be used to perform Bayesian inference in this setting as well.

\paragraph{Our contribution} In this paper, we present a new approach to posterior merging that leverages the recent advances in diffusion generative modelling, leading to a method that makes no distributional assumptions about the posteriors, such as Gaussianity, yet is able to scale well to complex and high dimensional posterior distributions where current methods for divide-and-conquer MCMC tend to perform poorly.

\section{Background on divide-and-conquer MCMC}\label{divideandconquerMCMCreview}

Divide-and-conquer MCMC methods aim to generate samples from the posterior distribution of parameter $\theta$, given samples from MCMC chains that are conditioned on subsets of the full dataset, $Y$. In this setting, $Y$ is partitioned into subsets $Y^\smallp{s}$, called \textit{shards}, which are often divided between multiple machines. MCMC chains are then run in parallel targeting the subposterior distributions $p^\smallp{s}$ conditioned on each shard of the data.  If the prior distribution $p(\theta)$ in each subposterior is scaled geometrically according to the number of shards $S$, then the subposteriors can be multiplied to obtain the full posterior distribution:
\begin{align}
    p^{\text{full}}(\theta) &:= p(\theta|Y) \propto \prod_s p(\theta)^{\frac{1}{S}} p(Y^\smallp{s}|\theta) = \prod_s p^\smallp{s}(\theta) \,.
\end{align}
Whilst this is a simple analytic relationship between $p^{\text{full}}$ and the $p^\smallp{s}$s, it is difficult to produce samples from $p^{\text{full}}$ given samples from each $p^\smallp{s}$. Current methods to do this come in two categories:

\paragraph{Approximately Gaussian} Since the dataset is typically very large in this setting, a natural approach is to appeal to the Bernstein-von Mises theorem and assume each subposterior distribution can be approximated by a Gaussian distribution. Several approximate methods in the literature are exact under this assumption, for example, parametric Gaussian estimation \citep{Neiswangeretal2014parallel} or methods based on transforming the subposterior samples to obtain approximate samples from the full posterior distribution \citep[e.g.][]{Scottetal2016consensus,Vyneretal2023swiss}. These methods are computationally efficient since no further MCMC sampling is required but are biased when the subposterior distributions are non-Gaussian, e.g. when they are skewed or multi-modal. However, non-Gaussianity is still quite common in the big data setting, since individual shards may not have enough data to form an informative posterior distribution for high-dimensional models.

\paragraph{Non-Gaussian approaches} Methods that are exact in the general setting do exist \citep[see][]{Chanetal2023fusion}, but are much slower than approximate methods. 
Methods based on non-parametric density estimation \citep[for example,][]{Neiswangeretal2014parallel,NemethSherlock2018} are asymptotically consistent in the number of MCMC samples and do not require the subposterior distributions to be approximately Gaussian. However, they struggle to scale to high-dimensional problems where a large number of samples of $\theta$ is required to approximate the posterior density since evaluating the approximation requires a computation over all of these MCMC samples.

\section{Methodology}\label{methodologysection}

\begin{figure}
\centering
\includegraphics[width=\textwidth]{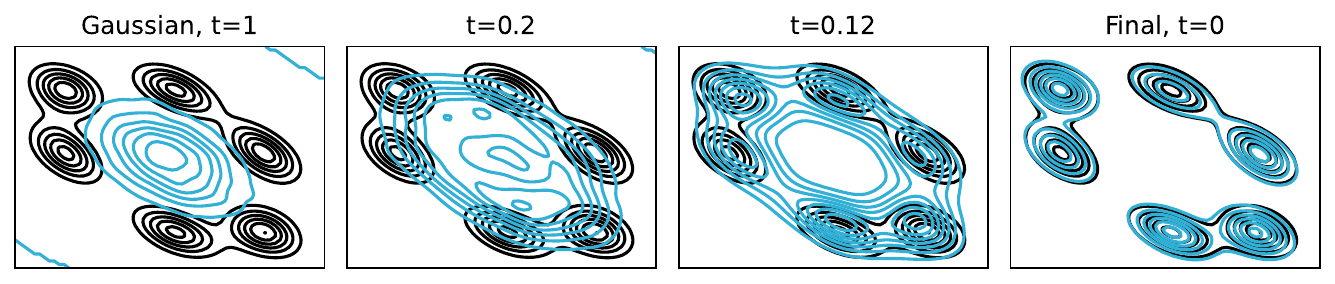}
\caption
{Annealed diffusion sampling in the mixture of Gaussians example. Full posterior in black.}\label{diffusionillustrationfigure}
\end{figure}

\begin{algorithm}
	\SetAlgoLined
	\vspace{2pt}
	\KwResult{Density estimates $\hat{p}_t(\theta,t)$ interpolating between a Gaussian and $\hat{p}^{\text{full}}(\theta)$}
	\textbf{Require:} Subposterior MCMC samples $\{\theta^\smallp{s}_{1:n_s}\}_{s=1}^S$ and score function evaluations $\{\nabla^\smallp{s}_{1:n_s}\}_{s=1}^S$\\
	\For{$s = 1, \ldots, S$}{
		\textbf{Affine transformation:}\\
        $\qquad \mu_s \leftarrow\text{mean}(\{\theta^\smallp{s}_{1:n_s}\})$;
        $V_s \leftarrow\text{matsqrt}(\text{cov}(\{\theta^\smallp{s}_{1:n_s}\}))$\\ \vspace{1pt}
        $\qquad \theta^\smallp{s}_i \leftarrow V_s^{-1}(\theta^\smallp{s}_i - \mu_s) \quad i\in\{1,\ldots,n_s\}$ \\\vspace{1pt}
        $\qquad \nabla^\smallp{s}_i \leftarrow V_s\nabla^\smallp{s}_i \quad i\in\{1,\ldots,n_s\}$
        \\	\vspace{1ex}
        \textbf{Train neural network parameters:}\\
        $\qquad \rho \leftarrow\argmin_{\rho} \; \EE_{t,\theta_t,\theta_0} [L^\smallp{s}(t,\theta_t,\theta_0;\rho)]$,\\\vspace{1pt}
        $\qquad$where $ t\sim U(0,1)$,  
        $\theta_t\sim p_{t|0}(\theta_t|\theta_0)$,   
        $\theta_0\sim\text{Categorical}(\{\theta^\smallp{s}_{1:n_s}\})$,        
        and 
        $\qquad L^\smallp{s}(t,\theta_t,\theta^\smallp{s}_i;\rho) = \| -\nabla_{\theta_t} E^\smallp{s}(\theta_t, t ; \rho) - \kappa_t \nabla_{\theta_t}\log p_{t|0}(\theta_t|\theta^\smallp{s}_i) - (1-\kappa_t)m(t)^{-1}\nabla^\smallp{s}_i \|_2^2$\\  
	}
	$\hat{p}_t(\theta,t) := \exp\left(-\sum_{s=1}^S E^\smallp{s}(V_s^{-1}(\theta-\mu_s), t; \rho^\smallp{s})\right)$
	\caption{Diffusion posterior approximation}\label{diffusionmergealgorithm}
\end{algorithm}

\paragraph{Overview} Our proposed approach to this problem involves training a neural network to approximate the unnormalised log-density of each subposterior distribution using diffusion generative modelling. We can then use the sum of the approximated log-densities in an MCMC sampler to sample from the full posterior distribution. Each diffusion model learns a sequence of densities interpolating between the subposterior distribution at time $t=0$, and a Gaussian approximation at $t=1$. This results in a sequence of approximations to the full posterior that interpolates between the parametric Gaussian approximation of \cite{Neiswangeretal2014parallel} and the final non-Gaussian posterior approximation. Figure \ref{diffusionillustrationfigure} illustrates the proposed diffusion posterior approximation algorithm on the Gaussian mixture example in Section \ref{mogsection}. Starting from a reference distribution, e.g. a Gaussian distribution, and implicitly smoothing the density estimates across time, the diffusion model can successfully learn complex, high-dimensional distributions. Using the sequence of densities in an annealed MCMC sampling procedure is also helpful for sampling from distributions where ordinary MCMC sampling struggles.
Algorithm \ref{diffusionmergealgorithm} gives an overview of the diffusion merging process. We retain the score function evaluations $\nabla^\smallp{s}_i := \nabla p^\smallp{s}(\theta^\smallp{s}_i)$ obtained while running the subposterior MCMC samplers in order to use a combination of the usual denoising score matching objective \citep{Vincent2011} and target score matching  \citep{debortoli2024target}, as described in Section \ref{tsmsection}.

\subsection{Diffusion generative modelling}\label{diffusionsreviewsection}

 \paragraph{Stochastic differential equations} Diffusion-based generative models \citep[e.g.][]{ho-etal-2020, Song-etal2021a}, work by defining a diffusion process that starts from a data distribution $p_0$ and adds noise to the data until the distribution converges to a known Gaussian prior distribution $p_1$. The time reversal of this process can be used to generate new samples from the data distribution. 
The noising process is a diffusion that is initialised at time $t=0$, which is the data distribution, and defined by an SDE of the form:
\begin{align}
    \d X_t = f(X_t,t)\d t + g(t)\d W_t \,.\label{nicesde}
\end{align}
Here, the drift term $f$ is usually linear in $X_t$ and controls the mean of the process, while the diffusion term $g$ controls the rate at which Gaussian noise is added. These are chosen so that the process converges to a Gaussian distribution as $t\rightarrow\infty$ regardless of the form of the true data density $p_{0}(x)$. By scaling the coefficients appropriately, it can typically be assumed that $X_t$ is approximately distributed according to the limiting distribution at time $t=1$. 
The most commonly used SDE in generative modelling is the variance-preserving (VP) SDE \citep{Song-etal2021a}, which is an inhomogenous Ornstein-Uhlenbeck process that converges to a standard Gaussian distribution:
\begin{align}
    \d X_t = -\frac{1}{2} \beta(t)X_t \,\d t + \sqrt{\beta(t)}\,\d W_t\,, \; \beta(t)>0 \,.\label{VPSDE}
\end{align}
If $\beta$ is chosen to be a linearly increasing function of $t$, this can be seen as a continuous version of the discrete sequence of noising kernels used by \citet{ho-etal-2020}, who perturb the data $x_0$ by sampling $x_i \sim \mathcal{N}(\sqrt{1 - \beta_i}x_{i-1},  \beta_i)$. 
\citet{Song-etal2021a} showed that the time reversal of an SDE of the form in \eqref{nicesde} has the following form, with time now running in reverse: 
\begin{align}
    \d X_t = \left[ f(X_t,t) - g(t)^2 \, \nabla \log p_t(X_t) \right]\d t + g(t) \d\tilde{W}_t  \, .\label{reverseSDE}
\end{align}
where $p_t(x)$ is the density of the noised distribution at time $t$, i.e. the marginal density of the SDE at time $t$ when initialised at $p_0$.

\paragraph{Score matching objective} In general, the density $p_t(x)$ is unknown, but we can approximate its score function $\nabla\log p_t$ using a parameteric model $\psi(x,t;\rho)$. The parameters $\rho$ of the function $\psi$ are estimated by minimising the denoising score matching objective \citep{Vincent2011}
\begin{align}
    L_{DSM}(\rho, t) &= \EE_{p_0(X_0)p_{t|0}(X_t|X_0)} \left[ \| \psi(x_t, t ; \rho ) - \nabla_{x_t} \log p_{t|0} (x_t|x_0) \|_2^2 \right] \,.\label{denoisingscorematchingobjective}
\end{align}
Note that this uses only the transition density $p_{t|0}$ of the diffusion process, which is simple to calculate for linear SDEs since $X_t - X_0$ has a Gaussian distribution whose parameters can be computed from the SDE coefficients.
The function $\psi$ is usually a single time-conditional neural network fit over all values of $t$ in $(0,1]$, so that it implicitly smooths score estimates across time. The full training objective is a weighted average of $L_{DSM}(\rho, t)$ across time, with $t$ uniformly sampled on $(0,1]$. 

\paragraph{Engery-based models} Since the score function of a distribution determines its density up to normalising constant, we can also use diffusion modelling for unnormalised density estimation, by parameterising the score function estimate as the gradient of a density function. This idea was suggested by \citet{salimans2021should} as a way of ensuring that the score function approximation is a conservative vector field. This is known as an energy-based parameterisation because we model an energy function $E(x,t;\rho)$ and approximate the unnormalised noised density $p_t$ by $\exp(-E(x,t;\rho))$. \cite{salimans2021should} proposed the parameterisation
\begin{align}
   E(x,t;\rho) &= \frac{1}{2s(t)} || x - \psi(x,t;\rho) ||^2_2\,,\label{energybasedparameterisation}
\end{align}
where $\psi(x,t;\rho):\RR^d\rightarrow\RR^d$ is a neural network and $s(t)^2$ is the variance of the noising kernel $p_{t|0}$. The gradient $-\nabla_x E(x,t;\rho)$ is substituted into the usual score matching objective in training, while $-E(x,t;\rho)$ models the log-density and can be used in MCMC sampling.

\subsection{Reparameterised stochastic differential equations}\label{normalisationsection}

Approximating the target posterior distribution using diffusion models can be challenging when the noise prior is significantly offset from the target distribution. Approximations to the target distribution trained with denoising score matching tend to poorly estimate the location and scale of the energy function. This could be addressed by normalising the dataset to have mean zero and unit covariance and will greatly improve the accuracy of the energy estimates. This normalisation is helpful for this problem in several different ways, easing neural network training and making the noise prior distributions in the diffusion models closer to their target distributions.

\paragraph{Transformed SDE} For any SDE with a standard Gaussian limiting distribution, this preprocessing is equivalent to using a modified SDE that converges to a Gaussian approximation to the target distribution, rather than a standard Gaussian. If $X_t$ evolves according to the linear SDE
\begin{align}
    \d X_t &= f(t)X_t \d t + g(t) \d W_t \,,\label{linSDE}
\end{align}
then by the It\^o formula \citep{Oksendal} the transformed process $\tilde{X}_t = AX_t + b$ has SDE
\begin{align}
    \d\tilde{X}_t &= f(t)(\tilde{X}_t - b)\d t + Ag(t) \d W_t \,,\label{shiftscaleSDE}
\end{align}
for $b\in\RR^d$ and $A\in\RR^{d\times d}$. By coupling, the limiting distribution of this SDE is $N(b+m,AS^2A^{\top})$, where $m$ and $S^2$ are the limiting mean and covariance of the original SDE. The score functions for the densities $p_t$ and $\tilde{p}_t$ of $X_t$ and $\tilde{X}_t,$ respectively, are related as follows:
\begin{align}
    \nabla\log \tilde{p}_t(x) &= \nabla\log \left\{ p_t( A^{-1}(x - b)) |A^{-1}| \right\} = \nabla\log p_t( A^{-1}(x - b))\,.\label{shiftedscore}
\end{align}
Thus, fitting a diffusion model to dataset $\tilde{X}$ using the transformed SDE \eqref{shiftscaleSDE} is equivalent to fitting a diffusion model to the transformed dataset $X = A^{-1}(\tilde{X} - b)$ using the original SDE \eqref{linSDE} and transforming the learned score function. 

\paragraph{SDE for product targets }When composing diffusion models, choosing different transformations for each component distribution is equivalent to using a different SDE for each diffusion model. This is not an issue, since we can still derive a sequence of noised densities that converges to the target product density as $t\rightarrow0$, by transforming the learned density functions $\hat{p}^\smallp{s}_t$:
\begin{align}
  \hat{p}_t(x,t) &\propto \prod_s \hat{p}^\smallp{s}_t(A^{-1}_s(x-b_s), t)\,.
\end{align}
In the divide-and-conquer setting, if the limiting distribution of SDE \eqref{linSDE} is a standard Gaussian, we propose choosing and $b_s$ and $A_s$ so that $b_s$ is the sample mean of subposterior $s$ and $A_sA^{\top}_s$ is its sample covariance. This makes the limiting distribution of SDE \eqref{shiftscaleSDE} a Gaussian approximation to that subposterior. The product of these limiting distributions, which is the noise prior $\hat{p}_1$ for the full posterior, is then equal to the Gaussian approximation to the full posterior suggested by \citet{Neiswangeretal2014parallel} (see Appendix \ref{sec:neiswanger-link}). Since the noise prior in diffusion models must be Gaussian, this choice is in a sense optimal for both the subposteriors and the full posterior, as it makes the noise priors as similar as possible to their respective target distributions. This means that our sequence of densities $\hat{p}_t$ interpolates between a Gaussian approximation to the full posterior and the learned non-Gaussian approximation. We follow \citet{Vyneretal2023swiss} in choosing $A$ to be the symmetric positive-definite square root of the sample covariance matrix $V$, $A = U\Lambda^{\frac{1}{2}} U^{\top}$, where $U$ and $\Lambda$ are the matrices of eigenvectors and eigenvalues, respectively, in the eigendecomposition $V = U\Lambda U^{\top}$.

This normalisation scheme makes it simpler to fit the neural network approximation to the density across time, since it effectively standardises the mean and variance of its inputs, which is known to improve fitting \citep{huangetal2023normalisation,karras2022elucidating}. When the normalised dataset $X$ is used with the variance preserving SDE, the mean and variance of $X_t$ are invariant over time. The original SDE \eqref{linSDE} has Gaussian marginal density $p_{0|t}$ with mean $m(t)x_0$ and covariance matrix $S(t)^2$, so that $X_t$ has mean $m(t)\EE(X_0)$ and variance $m(t)^2\Var(X_0) + S(t)^2$. 

\subsection{Alternative score matching objectives}\label{tsmsection}

\paragraph{Target score matching} Denoising score matching (DSM) often struggles to approximate the score function at low noise levels, since the variance of its score estimates explodes as $t\rightarrow0$. \citet{debortoli2024target} propose an alternative objective called target score matching (TSM) that has lower variance near time $t=0$, which can be used when it is possible to evaluate the unnoised log-density function $p_0$.
The target score matching loss proposed by \citet{debortoli2024target} is:
\begin{align}
    L_{TSM}(\theta,t) &= \EE_{X_0,X_t} \left[ \| \psi(x_t, t ; \theta ) - m(t)^{-1}\nabla \log p_{0} (x_0) \|_2^2 \right] \,,
\end{align}
which is designed so that estimates of the score of $p_t$ at $x_t$ are matched to a rescaling of the unnoised score of $p_0$ at $x_0$. 
The variance of Monte Carlo estimates in $L_{TSM}$ is low near $t=0$, but increases with $t$, exploding near $t=1$ for the variance preserving SDE, where $m(t)$ tends to 0. As such, \citet{debortoli2024target} suggest taking a convex combination of the regression targets of DSM and TSM, weighted in favour of TSM near $t=0$ and of DSM near $t=1$, yielding estimates of $\nabla \log p_t(x)$ that are well behaved across time. Following their suggestion, we minimise the objective function 
\begin{align}
    L(\rho,t) &= \EE_{X_0,X_t} \left[ \| \psi(x_t, t ; \rho ) - \kappa_t \nabla_{x_t}\log p_{t|0}(x_t|x_0) - (1-\kappa_t)m(t)^{-1}\nabla \log p_{0} (x_0) \|_2^2 \right], 
\end{align}
using a uniform weighting for $t$ and combination weights,
\begin{align}
    \kappa_t = \frac{s(t)^2}{s(t)^2 + m(t)^2\sigma^2_{\mathrm{data}}}\,,\label{tsmdsmweighting}
\end{align}
 which is optimal when the target has distribution $p_0 \sim N(0,\sigma^2_{\mathrm{data}}I_d)$. \citet{debortoli2024target} show that using this combined objective results in faster convergence to a mixture of Gaussians target distribution than using either DSM or TSM, which shows slower convergence than DSM since approximating the score well at times closer to 1 is important for SDE sampling.

\paragraph{Divide-and-conquer target score matching} In the case of gradient-based MCMC algorithms, such as the Hamiltonian Monte Carlo algorithm, the score function on the subposterior is calculated are stored as part of the MCMC training. This means that no additional subposterior evaluations are needed for training the target score matching objective. However, in order to use target score matching with normalised data, we must also rescale these score evaluations by $A$. This is because the training data for the neural network is the normalised dataset $X = A^{-1}(\tilde{X}-b)$ while the score function evaluations we have are for the density of the unnormalised $\tilde{X}$. We have $X_t = m(t)A^{-1}\tilde{X}_0 + W$, with $W\perp\!\!\!\perp \tilde{X}_0$, so the new regression target in TSM is $m(t)^{-1}A\,\nabla\log \tilde{p}_0(\tilde{x}_0)$, where $\nabla\log \tilde{p}_0(\tilde{x}_0)$ will have been computed and stored whilst running the MCMC sampler on the subposterior distribution. Using a normalised dataset in training means that its covariance is isotropic and therefore $\sigma^2_{\mathrm{data}}=1$ in weighting \eqref{tsmdsmweighting}.

\subsection{Model structure}\label{architecturesection}

All of our examples use a residual MLP, in line with \citet{Duetal2023RRR}. Details can be found in Appendix \ref{sec:network-architecture-detail}. 
The final output of the neural network $\psi$ was used to parameterise the energy function as follows:
\begin{align}
   E(x,t;\rho) &= - \frac{1}{2(m(t)^2 + s(t)^2)} || x - \psi(x,s(t);\rho) ||^2_2\,,
\end{align}
This is based on the parameterisation proposed by \citet{salimans2021should} described in Equation \eqref{energybasedparameterisation}, with the output scaling adjusted to prevent the energy function from degenerating as $t \rightarrow 0$. This choice was inspired by the similarity of this parameterisation to the Gaussian density function -- the variance of $X_t$ will be $m(t)^2\Var(X_0) + s(t)^2I_d$, which is $(m(t)^2 + s(t)^2)I_d$ if the data is normalised as suggested in Section \ref{normalisationsection}. This means that  a constant output of $0$ will match the true energy function for a Gaussian target when a normalised dataset is used. Regardless of the target distribution, $E$ will tend to the energy function for the noise prior as $t\rightarrow\infty$ as long as the output of $\psi$ tends to 0.

\subsection{Sampling from compositions of diffusion models}\label{diffusioncompsection}

In order to use diffusion modelling in divide-and-conquer MCMC, we need to be able to combine different models to generate samples from the product of their target distributions. A naive approach would be to add the component score functions $\nabla \log p^\smallp{s}_t$ together to obtain the score of the product of the $p^\smallp{s}_t$. 
At time $t=0,$ this is exactly equal to the score of the target product distribution $p^{\text{full}}_t$. However, this relationship does not hold for $t>0$ since we cannot interchange noising with multiplication of densities.  
Indeed, simply substituting the score sum into the reverse SDE fails to generate samples from the correct distribution. This fails even for  Gaussian targets, where for $t>0$, the score sum estimate corresponds to a Gaussian distribution with a different mean and variance to $p^{\text{full}}_t$ (see Appendix \ref{sec:score-sum-failure}). 

As we have an enery-based estimate for each noised sub-posterior, our solution is to use an annealed MCMC sampling procedure \citep{GeffnerSBI,Duetal2023RRR}, since the sequence of densities, $\hat{p}_t$, obtained by multiplying the noised component densities together interpolates smoothly between a tractable noise prior and the target product distribution. We can then transport samples from the new prior to the target by starting with a sample of size $N$ from the prior at $t=1$ and then iteratively using a fixed number of unadjusted MCMC updates to target $\hat{p}_t$ for a sequence of predetermined timepoints $1>t_2>\ldots>t_n=0$ (see Appendix \ref{sec:aMCMC-sampling}). 
Using an energy based parameterisation  enables the use of a Metropolis-Hastings adjusted sampler in this procedure, which has been shown to improve results for compositional generation \citep{Duetal2023RRR}. 

Using the parameterisation described in Section \ref{architecturesection} instead of the choice given in  \citet{salimans2021should}, we can obtain an estimate for the unnoised target density at time $t=0$. We found in our experiments that in some cases this could be used within an ordinary MCMC sampler to generate samples from the full posterior with equivalent accuracy to the annealed sampling method. For particularly challenging distributions, e.g. multimodal distributions, the annealed sampling procedure had better mixing and allowed accurate recovery of the mode weights.

\section{Related work}
\label{sec:related-work}

\paragraph{Normalising flows} The most similar method in the literature is that of \citet{mesquita2020nvp}, who approximate the subposteriors with a discrete normalising flow, i.e. a sequence of invertible neural network transformations mapping a reference distribution onto the target. Since their work, diffusion-based modelling has emerged as an alternative to normalising flows that is more efficient to train and performs better on density estimation \citep{Song-etal2021a}. In addition, our formulation enables the use of a wider class of non-invertible neural network architectures and parameterises a density estimate that can be evaluated without using the change-of-variables formula.

\paragraph{Neural density estimation} Score matching methods have previously been used in neural density estimation by \citet{saremi2018deep}, who use denoising score matching to train a neural network approximation to a kernel density estimate. \citet{SongErmon2019} note that this estimator can be difficult to use in MCMC sampling as fixing a small bandwidth  produces poor estimates in low density areas, causing poor mixing of the MCMC sampler and a failure to recover mode weights in multimodal distributions. This would cause issues in divide-and-conquer MCMC since the subposterior distributions often have poor overlap with the full posterior, so accuracy in low density areas is important. 
The use of diffusion models to learn an energy function approximation was proposed by \cite{salimans2021should} and developed further by \cite{Duetal2023RRR} in order to sample from compositions of diffusion image generation models.

\section{Experiments}\label{experimentssection}

The experiments in this section were chosen to demonstrate that diffusion models can be used to accurately recover the full posterior distribution in divide-and-conquer problems where other merging methods struggle, in particular where the subposteriors have poor overlap or are significantly non-Gaussian.
We compare to the following methods in the literature:
\begin{itemize}
    \item Consensus Monte Carlo (CMC) \citep{Scottetal2016consensus}, where a weighted average of samples from each shard is taken.
    \item Sub-posteriors with inflation, scaling and shifting (SwISS) \citep{Vyneretal2023swiss}, where subposteriors are transformed with an affine transformation to approximate the full posterior.
    \item Gaussian parametric density estimation \citep{Neiswangeretal2014parallel}, which is identical to the noise prior used in the diffusion approximation.
    \item Semiparametric density estimation \citep{Neiswangeretal2014parallel}, which is a product of the Gaussian estimate and a nonparametric correction factor that scales better to high dimensional problems than pure kernel density estimation. This was implemented with the R package parallelMCMCcombine \citep{Miroshnikov2014parallelmcmccombine}.
    \item Gaussian process approximation to the log density  \citep{NemethSherlock2018}, using the GPJax python package \citep{Pinder2022gpjax}. 
\end{itemize}
The merging methods were compared numerically to samples generated from the full posterior using three sample based discrepancy metrics: Mahalanobis distance (Mah), integrated absolute distance (IAD), and mean absolute skew deviation (Skew). 
Full experimental details as well as the numerical comparison for the toy examples can be found in Appendix \ref{experimentdetailssection}. 
Code to reproduce our experiments can be found at \url{https://github.com/ctrojan/DiffusionDnC}. 
We report training times for the methods requiring an optimisation phase, and sampling times for the methods requiring additional MCMC sampling.

The diffusion approximations used the variance preserving SDE. The neural network architecture was the same for each experiment, with the exception of the input and output layers which must have the same size as the dimension of the target distribution. The number of training epochs was also chosen so that the number of training updates was the same for each experiment. This was done to highlight the fact that the diffusion models did not require additional hyperparameter tuning to perform well across different experiments. The neural network training on each shard can be done in parallel, so while training time makes up the majority of the execution time, it does not depend on the number of shards. This made the execution time of the diffusion merging algorithm very similar across experiments, regardless of the complexity of the target distribution, the number of shards, or the length of the MCMC chains sampled from each subposterior. 

\begin{figure}
\centering
\includegraphics[width=\textwidth]{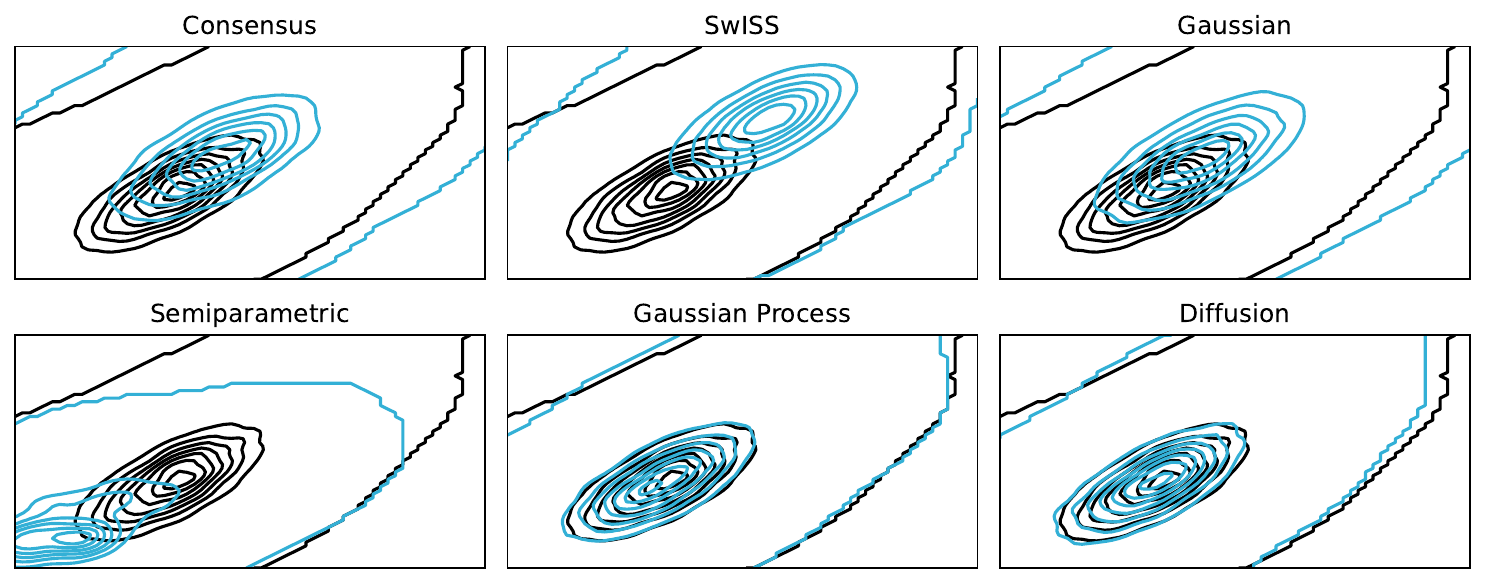}
\caption
{Merged posterior contour plots for the toy logistic regression example.}\label{toylogregmergescomparisonfigure}
\end{figure}

\subsection{Toy logistic regression}
Our first example is a synthetic logistic regression dataset,
with a 1-dimensional covariate $x\sim N(0.5,1)$. 
The true value of the parameter of interest is $\theta=(-3,-3)$, leading to a low positive rate of around 0.1. 1000 datapoints were generated  and split across 15 shards, creating a moderately challenging merging problem as the number of positive examples on each shard varies considerably, so the subposteriors are both non-Gaussian and very dissimilar.
See Figure \ref{toylogregmergescomparisonfigure} for the posterior contour plots for each method -- note that the location is difficult to estimate without using density estimation, likely because the subposteriors are skewed so assumptions of Gaussianity do not hold. Only the Gaussian process and diffusion approximations recovered the true posterior with reasonable accuracy.

\subsection{Toy mixture of Gaussians}\label{mogsection}

\begin{figure}
\centering
\begin{subfigure}[b]{.21\textwidth}\centering
\includegraphics[height=2.5cm]{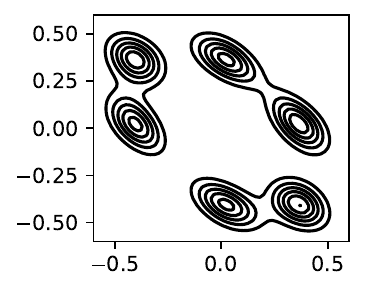}
\caption{Full posterior}
\end{subfigure}
\begin{subfigure}[b]{.78\textwidth}\centering
\includegraphics[height=2.5cm]{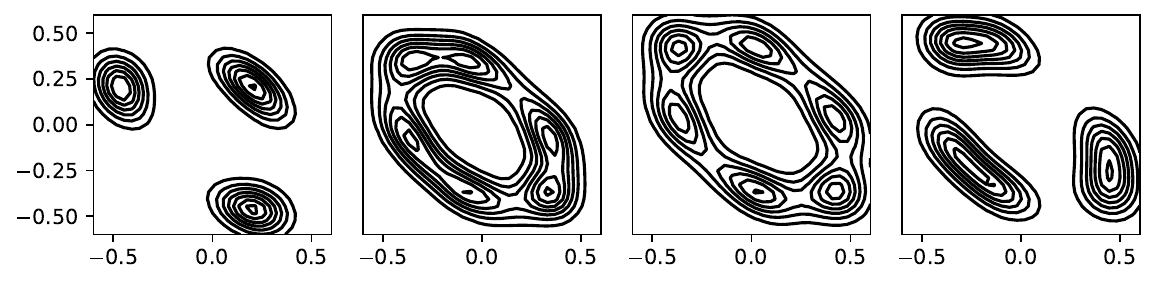}
\caption{Subposteriors}
\end{subfigure}
\caption
{Mixture of Gaussians posterior contour plots for $\theta_1$ and $\theta_2$.}\label{mogpostsfigure}
\end{figure}

\begin{figure}
\centering
\includegraphics[width=\textwidth]{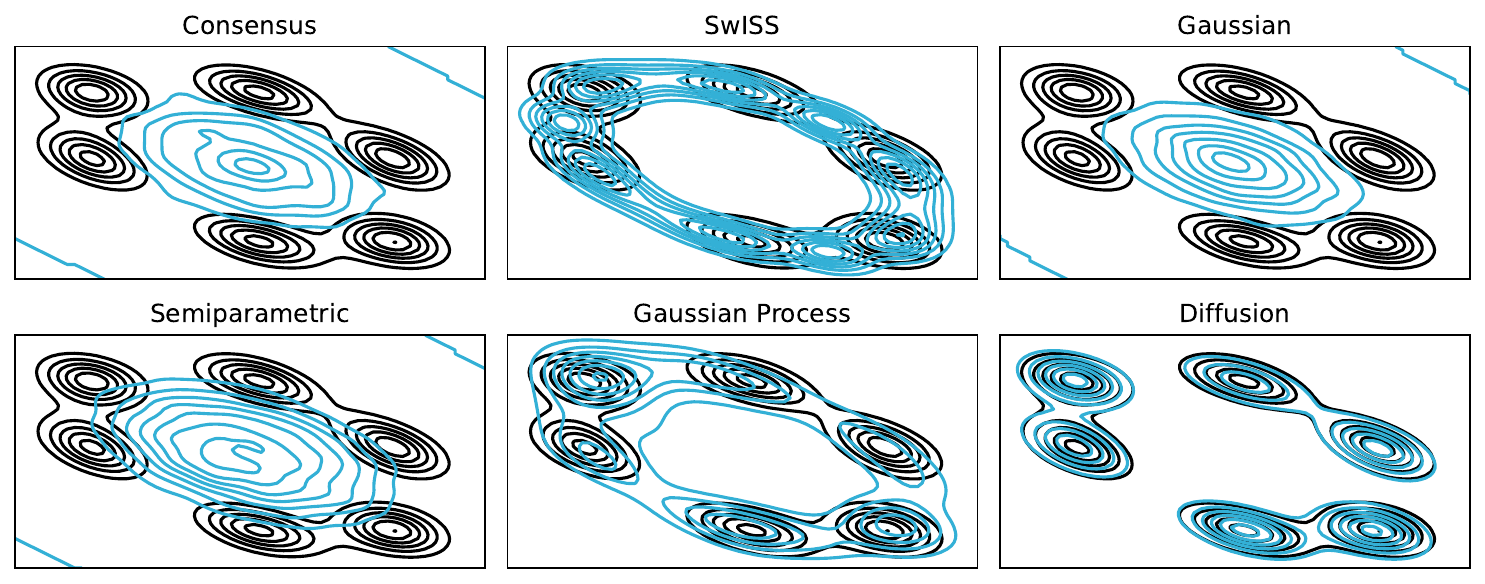}
\caption
{Merged posterior contour plots for first two parameters in the mixture of Gaussians example.}\label{mogmergescomparisonfigure}
\end{figure}

In this example, the data was drawn from a 1D mixture of 3 Gaussians, $x \sim \frac{1}{3}N(\theta_1,0.2)+\frac{1}{3}N(\theta_2,0.2)+\frac{1}{3}N(\theta_3,0.2)$ with $\theta = (0.4,0,-0.4)$. This gives a posterior distribution with 6 modes since the likelihood is invariant to label switching of the $\theta_i$. The generated dataset was of size 2000, and was split across 4 shards. See Figure \ref{mogpostsfigure} for an illustration of the full posterior and subposteriors -- note that the subposterior modes appear in differing locations, making it difficult to recover the structure of the full posterior.
See Figure \ref{mogmergescomparisonfigure} for a comparison of the merged posterior contour plots for $\theta_1$ and $\theta_2$. The diffusion approximation was the only method to accurately recover the full posterior's mode locations and weights.

\subsection{Robust linear regression on the combined cycle power plant dataset}

Inspired by \cite{Chanetal2023fusion}, we consider a robust linear regression example on the combined cycle power plant dataset \citep{combined_cycle_power_plant_294}. This dataset consists of 9568 hourly observations of 4 features as well as the net hourly electrical output of the power plant, which is the regression target. The model consists of a linear fit to the data with $t$-distributed errors to increase robustness to outliers. We sample from the joint posterior distribution of the regression coefficients $\beta\in\RR^5$ and noise scale $\sigma\in\RR_{>0}$, fixing the degrees of freedom of the noise distribution to 5.
We split the data randomly across 8 shards, reporting results in Table \ref{powerplantmergestable} as an average over 5 splits of the dataset, with standard deviations in parentheses. In this example, the full marginal distribution of $\sigma$ is very difficult to estimate from the subposteriors as is is highly skewed and has poor overlap with the subposteriors. No method succeeded at accurately recovering this marginal, but the SwISS and diffusion approximations were closest to the true location and had the best accuracy overall.

\begin{table}
\caption{Power plant discrepancies and average wall clock execution time.}
\label{powerplantmergestable}
\centering
\begin{tabular}{lccccc}
\toprule
\textbf{Method}  & \textbf{Mah} & \textbf{IAD} & \textbf{Skew} &  \textbf{Training} & \textbf{Sampling} \\ \midrule
Consensus & 6.25 (0.04) & \textbf{0.21} (0.02) & \textbf{0.06} (0.00) & - & -\\
SwISS & \textbf{4.14} (0.03) & \textbf{0.21} (0.02) & \textbf{0.06} (0.00) & - & -\\
Gaussian & 6.25 (0.04) & \textbf{0.21} (0.02) & 0.14 (0.00) & - & -\\ \midrule
Semiparametric & 5.74 (0.89) & 0.90 (0.08) & 2.88 (1.38) & - & 255s \\
Gaussian process & 7.58 (0.07) & 0.28 (0.01) & 0.14 (0.01) & 82s & 1743s\\
Diffusion & \textbf{4.14} (0.04) & \textbf{0.21} (0.02) & 0.07 (0.01) & 100s & 5s\\
\bottomrule
\end{tabular}
\end{table} 

\subsection{Logistic regression on the spambase dataset}

We fit a logistic regression to the spambase dataset \citep{spambase_94}, which consists of 4600 e-mails classified as spam or not spam, summarised by 57-dimensional vectors of word and character frequencies. The parameter space is relatively high dimensional ($d=58$), creating a challenging merging problem since the subposteriors vary in shape and have poor overlap, as well as requiring a large number of samples to summarise each distribution. We split the data randomly across 4 shards. Results are reported in Table \ref{spammergestable} as an average over 5 splits, with standard deviations in parentheses.
The diffusion merge significantly outperformed the other algorithms on Mahalanobis distance and IAD, showing that the location and scale of the full posterior distribution were recovered more accurately. However, the skew deviation was higher than for the SwISS algorithm due to underestimation of the skew on the subposteriors. 

\begin{table}
\caption{Spambase discrepancies and average wall clock execution time.}
\label{spammergestable}
\centering
\begin{tabular}{lccccc}
\toprule
\textbf{Method}  & \textbf{Mah} & \textbf{IAD} & \textbf{Skew} &  \textbf{Training} & \textbf{Sampling} \\ \midrule
Consensus & 6.00 (0.76) & 0.24 (0.02) & 0.26 (0.01) & - & -\\
SwISS & 7.04 (0.64) & 0.26 (0.02) & \textbf{0.22} (0.01) & - & -\\
Gaussian & 6.02 (0.77) & 0.24 (0.02) & 0.37 (0.00) & - & -\\ \midrule
Semiparametric & 6.10 (0.35) & 0.29 (0.02) & 0.42 (0.04) & - & 304s\\
Gaussian process & 6.25 (0.96) & 0.29 (0.03) & 0.37 (0.00) & 69s & 1246s \\
Diffusion & \textbf{4.54} (0.79) & \textbf{0.17 }(0.02) & 0.26 (0.01) & 149s & 4s\\
\bottomrule
\end{tabular}
\end{table}

\section{Discussion}

In this paper we proposed the use of diffusion generative modelling to  merge MCMC samples generated in parallel from disjoint subsets of the full dataset. The resulting method is embarrassingly parallel, with the exception of the final merging step where a new MCMC run is performed using the diffusion approximations. Our method outperformed existing merging algorithms in the literature on complex and high dimensional posterior distributions. It is also more computationally efficient on complex problems than existing density estimation approaches. This is because the MCMC sampling stage is very efficient -- the density approximation is cheap to evaluate, with a cost that is independent of the number of samples used to train it. The majority of the computational cost comes from the training time of the neural networks, which scales well to larger problems and can be done in parallel.

\paragraph{Limitations} The main limitation to our approach is that it is more computationally costly than methods which do not require optimisation or further MCMC sampling, such as consensus Monte Carlo and SwISS. In cases where the simplifying assumptions made by these methods fail, however, it can be used to recover the full posterior distribution with greater accuracy at a moderate computational cost.
\FloatBarrier

\begin{ack}
CT acknowledges the support of the EPSRC-funded EP/S022252/1 Centre for Doctoral Training in Statistics and Operational Research in Partnership with Industry (STOR-i); PF was supported by EPSRC grants EP/Y028783/1, EP/R034710/1 and EP/R018561/1; CN was supported by EPSRC grants EP/Y028783/1 and EP/V022636/1. 
\end{ack}

{
\small
\bibliographystyle{abbrvnat}
\bibliography{bibliography}

}

\FloatBarrier
\appendix
\section{Experimental details}\label{experimentdetailssection}

Experiments were run using Python 3 and the JAX package \citep{jax2018github}. Real datasets were provided by the UCI machine learning repository \citep{ucimlrepo} via the ucimlrepo Python package. Both the spambase and combined cycle power plant datasets are released under a CC BY 4.0 licence. Experiments were run on a Linux virtual machine with Ubuntu 22.04.1 LTS on CPU, which was an Intel® Xeon(R) Gold 6248R CPU @ 3.00GHz × 4.

\subsection{Discrepancy metrics}

The merging methods were compared numerically to samples generated from the full posterior using the following sample based discrepancy metrics:

\paragraph{Mahalanobis distance} $d_{Mah} = \sqrt{(\mu_a - \mu_f)^{\top}\,V_f^{-1}\,(\mu_a - \mu_f)}$, where $\mu_a$ and $\mu_f$ are the means of the approximate and exact samples respectively and $V_f$ is the sample covariance matrix of the exact samples. This is useful for assessing how close the location of the approximate samples is to that of the true distribution, accounting for different scales in different directions.
    
\paragraph{Integrated absolute distance} $d_{IAD} = \frac{1}{2d} \sum_{i=1}^d \int_{\Theta_i} | \hat{\pi}_{a,i}(\theta_i) - \hat{\pi}_{f,i}(\theta_i) | \,\d\theta_i$ . Here, $\hat{\pi}_{.,i}$ is a kernel density estimate of the marginal density of dimension $i$ based on the approximate ($\hat{\pi}_{a,i}$) or exact ($\hat{\pi}_{f,i}$) samples. Integration range was chosen by computing the union of the 5-sigma intervals centered on the mean for both sets of samples, which was sufficient to accurately obtain the number of decimal places reported. This measures discrepancy on the 1D marginals.
    
\paragraph{Mean absolute skew deviation} $d_{\text{skew}} = \frac{1}{d}\sum_{i=1}^d | \hat{\gamma}_f^{(i)} - \hat{\gamma}_a^{(i)}|$, where $\hat{\gamma}^{(i)}_.$ is a sample approximation of the skewness of the 1D marginal density of component $i$. Here, $\hat{\gamma}^{(i)}=\frac{1}{n}\sum_{j=1}^n [(\theta_i - \mu_i)/\sigma_i]^3$, $\mu_i$ and $\sigma_i$ being the sample mean and standard deviation of component $i$. This is a measure of similarity in shape that is independent of differences in location and scale between the distributions.

\subsection{MCMC sampling}
MCMC inference was carried out with the BlackJAX Python package \citep{cabezas2024blackjax}. The NUTS algorithm was used, with step sizes tuned to achieve an acceptance rate close to 80\%. In the toy examples, samplers were initialised at an area of high posterior density and 10\,000 samples were drawn from each subposterior after a burn in of 100.

Since the evaluation time of the Gaussian process approximation scales quadratically with the number of inducing points, we follow \citet{NemethSherlock2018} in thinning the MCMC chains before use. In each example the subposterior chains were thinned to 1000 samples.

\subsection{Diffusion sampling}\label{sec:aMCMC-sampling}

For sampling from the diffusion approximations, we used the learned density approximation at a fixed time $t=0$ in the NUTS algorithm, with the exception of the mixture of Gaussians experiment where the annealed sampling procedure of \cite{Duetal2023RRR} was used. This algorithm is described in more detail in Algorithm \ref{aMCMCalgorithm}.

\begin{algorithm}
	\SetAlgoLined
	\vspace{2pt}
	\KwResult{Sample $\theta$ from the target distribution}
	\textbf{Require:} Sequence $p_t$ of target densities,
 MCMC update function and\\ $\qquad\qquad$ its hyperparameters $\eta$, number of steps $n_{outer}$, $n_{inner}$\\
 $t \leftarrow 1$\\
 $\theta_1 \sim p_1$\\
	\For{$i \in \{1,\ldots,n_{outer}\}$}{
        $t \leftarrow t - 1/n_{outer}$\\
        $\theta_t \leftarrow \theta_{t-1}$\\
        \vspace{2pt}
    	\For{$j \in \{1,\ldots,n_{inner}\}$}{
            $\theta_t \leftarrow \mbox{MCMC\_update}(\theta_t; p_t, \eta)$\\
    	}      
	}
 $\theta \leftarrow \theta_t$
	\caption{Annealed MCMC sampling}\label{aMCMCalgorithm}
\end{algorithm}

\subsection{Details for individual experiments}

\begin{figure}
\centering
\includegraphics[width=\textwidth]{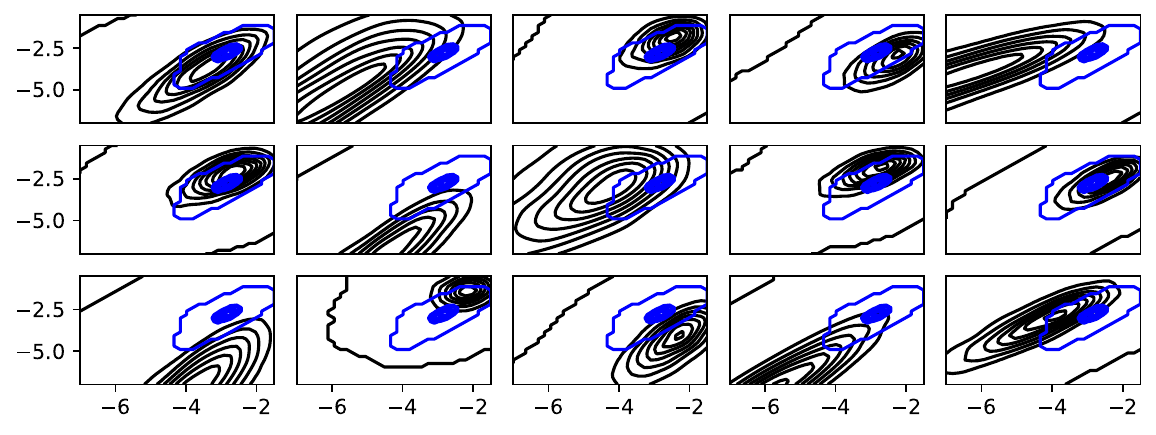}
\caption{\footnotesize Subposterior contours for the logistic regression with full posterior in blue.}\label{toylogreg15subposts}
\end{figure}

\paragraph{Toy logistic regression}
In this example, the dataset consisted of a 1-dimensional covariate $x_i\sim N(0.5,1)$ with binary class labels generated as $y_i\sim \text{Bernoulli}(\text{sigmoid}(\beta_1 +\beta_2x_i))$. The true value of the parameter of interest was $\beta=(-3,-3)$. A normal prior of $N(0,5)$ was placed over all parameters. An illustration of the subposteriors can be found in Figure \ref{toylogreg15subposts} - note that they are very skewed and differ substantially in location and scale. Most of them have poor overlap with the full posterior (overlaid in blue) which is very concentrated around the true parameter value. The numerical results of the algorithms are reported in Table \ref{toylogregmergestable}.

\begin{table}
\centering
\caption{Toy logistic regression discrepancies and wall clock time.}
\label{toylogregmergestable}
\begin{tabular}{lccccc}
\toprule
\textbf{Method}  & \textbf{Mah} & \textbf{IAD} & \textbf{Skew} &  \textbf{Training time} & \textbf{Sampling time} \\ \midrule
Consensus & 1.28 & 0.42 & 0.02 &-&- \\
SwISS & 2.57 & 0.75 & 0.03 &-&- \\
Gaussian & 1.28 & 0.41 & 0.17 &-&- \\ \midrule
Semiparametric & 2.07 & 0.63 & 0.74 &-& 65s \\
Gaussian process & \textbf{0.08} & 0.04 & 0.22 & 36s & 2766s \\
Diffusion & \textbf{0.08} & \textbf{0.03} & \textbf{0.01} & 99s & 8s \\
\bottomrule
\end{tabular}
\end{table}

\paragraph{Toy Gaussian mixture}
A standard Normal prior placed over all parameters. To ensure good mixing between modes when sampling from the full posterior and subposteriors, MCMC samplers randomly permuted the $\theta_i$ at each step as in \citet{Neiswangeretal2014parallel}, since this is a move that leaves the posterior density invariant. Label switching was not used for sampling from the merged distributions in order to give an accurate representation of the fitted density. Instead, the annealed sampling procedure was used in the diffusion, and 10 independent chains were run for the Gaussian process. The annealed sampling for the diffusion model used HMC rather than NUTS since it allows the update steps to be vectorised efficiently. The number of leapfrog steps was set at 3 and a fixed step size was used across all times. 300 evenly spaced time points were used, and 1 MCMC step was performed at each time point. Numerical results are reported in Table \ref{mogmergestable}.

\begin{table}
\centering
\caption{Mixture of Gaussians discrepancies and wall clock time.}\label{mogmergestable}
\begin{tabular}{lccccc}
\toprule
\textbf{Method}  & \textbf{Mah} & \textbf{IAD} & \textbf{Skew} &  \textbf{Training time} & \textbf{Sampling time} \\ \midrule
Consensus & 0.14 & 0.53 & 0.16 & - & -\\
SwISS & 0.16 & 0.25 & 0.14 & - & -\\
Gaussian & 0.13 & 0.55 & 0.14 & - & -\\ \midrule
Semiparametric & 0.14 & 0.52 & \textbf{0.12} & - & 24s \\
Gaussian process & 0.19 & 0.24 & 0.21 & 35s & 3412s \\
Diffusion & \textbf{0.11} & \textbf{0.04} & \textbf{0.12} & 98s & 24s\\
\bottomrule
\end{tabular}
\end{table}

\paragraph{Power plant robust regression}
Priors of $N(0,10^2)$ were placed on the regression coefficients, and for the noise standard deviation a chi-squared prior $\chi(1)$ with scale parameter 10 was used. MCMC samplers were initialised at the mean of the prior distribution. 50\,000 samples were generated from each posterior after a burn in of 100. For the merged distributions, the GP and diffusion samplers were initialised at the mode of the Gaussian approximation to the full posterior.

\paragraph{Spambase logistic regression} 
Priors of $N(0,5)$ were placed on all parameters. The full posterior and subposterior MCMC samplers were initialised at areas of high posterior density by using 20 epochs of the ADAM optimiser with a learning rate of $10^{-2}$ to approximately find the MAP parameter estimate. 50\,000 samples were generated from each posterior after a burn in of 100. For the merged distributions, the GP and diffusion samplers were initialised at the mode of the Gaussian approximation to the full posterior. Using the inverse of its covariance matrix as the inverse mass matrix in NUTS sampling greatly reduced the number of leapfrog steps required to sample from the full posterior approximation.

\subsection{Neural network architecture}\label{sec:network-architecture-detail}

The neural network used was a residual MLP, implemented with the Flax Python package \citep{flax2020github}. The noise standard deviation $s(t)$ was concatenated to the input of each layer. In this architecture, each hidden layer has the same dimension, with the exception of the output layer which has the same dimension as $x$. After the first layer, the hidden layers are organised in blocks of two with skip connections that add the input of the block to the output of its hidden layers. The size of the network was kept the same across experiments, with the exception of the output layer which must have the same dimension as the target distribution. A neural network with 1 residual block and the hidden layer dimension 32 was sufficient for the examples considered. The activation function for the hidden layers was $\text{SiLU}(x)=x\,\text{sigmoid}(x)$, a smooth alternative to the ReLU activation function \citep{ramachandran2017searching}.

\subsection{Training}

The Adam optimiser \citep{Adam} as implemented by the Optax Python package \citep{deepmind2020jax} was used for model training, with a batch size of 32 and default hyperparameters. The Gaussian process parameters were fit for 200 epochs. For the diffusion models, 500 epochs of training were used for the toy examples, and 100 for the power plant and spambase examples.

\section{Scalability to very high-dimensional problems}
Here, we report the results of an additional high-dimensional synthetic logistic regression experiment to show that our method also scales well to dimensions higher that those considered in the main paper. In this experiment, the dataset consisted of 1000 realisations of a 99-dimensional covariate $x$ drawn from a standard Normal distribution, with binary class labels generated as $y_i\sim \text{Bernoulli}(\text{sigmoid}(\beta_0 +\beta^{\top}x_i))$. The true value of $\beta_0$ was -5, and the remaining $\beta_i$ were drawn from a standard Normal distribution. A prior of $N(0,5)$ was placed over all parameters. The dataset was split across 4 shards and 50\,000 samples were drawn from all posterior distributions after a burn in of 100. Experimental details were otherwise the same as in the other experiments.

This was repeated for 5 simulated datasets and the numerical results are reported in Table \ref{table:bigtoylogreg} as an average over the 5 runs with standard deviations in brackets. SwISS outperformed all other methods here as the subposteriors were similarly shaped and had low skew. Note that with $d=100$ here, the non- and semi- parametric density estimation approaches had significantly worse performance than the other methods, while their execution time increased significantly compared to the lower-dimensional experiments. In comparison, the diffusion method was outperformed only by SwISS and its execution time was not much higher than in the other experiments.

\begin{table}
\caption{High-dimensional logistic regression discrepancies and average execution time.}
\label{table:bigtoylogreg}
\centering
\begin{tabular}{lccccc}
\toprule
\textbf{Method}  & \textbf{Mah} & \textbf{IAD} & \textbf{Skew} &  \textbf{Training} & \textbf{Sampling} \\ \midrule
Consensus & 4.73 (0.17) & 0.25 (0.00) & 0.02 (0.00) & - & -\\
SwISS & \textbf{3.36} (0.17) & \textbf{0.17} (0.01) & \textbf{0.01} (0.00) & - & -\\
Gaussian & 4.73 (0.17) & 0.25 (0.00) & 0.02 (0.00) & - & -\\\midrule
Semiparametric & 6.94 (0.25) & 0.31 (0.01) & 0.03 (0.00) & - & 733s \\
Gaussian process & 7.41 (0.17) & 0.32 (0.02) & 0.03 (0.00) & 92s & 12025s \\
Diffusion & 4.02 (0.17) & 0.21 (0.00) & 0.02 (0.00) & 189s & 10s\\
\bottomrule
\end{tabular}
\end{table}

\section{Link between diffusion formulation and parametric Gaussian approximation}\label{sec:neiswanger-link}

In the proposed merging algorithm, we obtain a sequence of density estimates for each subposterior that interpolates between a Gaussian prior distribution $\hat{p}^\smallp{s}_1$ and the final non-Gaussian diffusion approximation $\hat{p}^\smallp{s}_0$. Using the reparameterisation in Section \ref{normalisationsection}, the Gaussian priors are $N(\mu_s,V_s)$, where $\mu_s$ and $V_s$ are the sample mean and covariance respectively of the samples $\{\theta^\smallp{s}\}$ generated from shard $s$.

The Gaussian prior distribution $\hat{p}_1$ for the full posterior is the product of those of the subposteriors, and has the form $N(\mu,V)$ where:
\begin{align}
 \mu = V\sum_s V_s^{-1}\mu_s \quad \mbox{and} \quad V = \Big[\sum_s V_s^{-1}\Big]^{-1}\,.
\end{align}
This is exactly the parametric Gaussian approximation of \cite{Neiswangeretal2014parallel}.

\section{Failure of SDE sampling for compositions of diffusions}\label{sec:score-sum-failure}

By adding the component score functions $\nabla \log p^\smallp{s}_t$ together, we obtain the the score of the product distribution
\begin{align}
    \Tilde{p}_t(\theta_t) &\propto \prod_s p^\smallp{s}_t(\theta_t)\,,
\end{align}
which at time $t=0$ is exactly equal to the score of the target product distribution $p^{\text{full}}$. However, this relationship does not hold for $t>0$ since we cannot interchange the order of adding noise to a distribution with multiplying densities together:
\begin{align}
    p^{(\text{full})}_t(\theta_t) 
    &= \int_{\Theta} p_{t|0}(\theta_t|\theta_0) \; p^{\text{full}}(\theta_0) \, \d \theta_0  
    \propto \int_{\Theta} p_{t|0}(\theta_t|\theta_0)\prod_s p^\smallp{s}(\theta_0) \, \d\theta_0\\
    &\not\equiv  \Tilde{p}_t(\theta_t)
    \propto \prod_s \int_{\Theta} p_{t|0}(\theta_t|\theta_0) \; p^\smallp{s}(\theta_0) \, \d\theta_0 \, .\nonumber
\end{align}
This relationship fails even when the target distributions $p_0^\smallp{s}$ are Gaussian. For linear SDEs the marginal distribution $p_{t|0}$ is $N(m(x_0,t), S(t)^2)$, where $m$ is linear in $x_0$, i.e. $m(x,t) = m_1(t)x + m_2(t)$. So, when the target $p_0$ is Gaussian with parameters $\mu$ and $V$, $p_t$ will also be Gaussian and we can compute its parameters as $m(\mu, t)$ and $m_1(t)^2V + S(t)^2$, since $m$ is linear and $x_t$ can be expressed as $m(x_0,t) + S(t)Z$ where $Z$ is a standard normal.

So, assuming $p_0^\smallp{s}=N(\mu_s,V_s)$, we can obtain $p^{\text{full}}_t$ by taking a product of Gaussian densities and then adding noise, i.e.,
\begin{align}
    p^{\text{full}}_t &\propto \phi(m(\mu, t), m_1(t)^2V + S(t)^2)\,, \ \mbox{where} \\ \mu &= V\sum_s V_s^{-1}\mu_s\,, \quad \mbox{and} \quad V = \Big[\sum_s V_s^{-1}\Big]^{-1}\,.
\end{align}
However, $\tilde{p}_t$ is a product of the noised densities $p_t^\smallp{s}$ and will be Gaussian with parameters:
\begin{align}
    \tilde{\mu} = \tilde{V}\sum_s [m_1^2(t)V_s+S(t)^2]^{-1}m(\mu_s)\,, \quad \tilde{V} = \Big[\sum_s [m_1(t)^2V_s + S(t)^2]^{-1}\Big]^{-1} \,. 
\end{align}
So, unless the $\mu_s$ are zero, the mean of $\tilde{p}_t$ will typically be different to that of $p^{\text{full}}_t$. Its variance will also be incorrect - even if all of the $V_s$ are equal, we obtain $\tilde{V} = m_1(t)^2V + \frac{1}{S}S(t)^2$, so taking the product in this way effectively shrinks the noise added by the diffusion. 

Hence, while the score sum does get closer to the desired score function $\nabla\log p^{(\text{full})}_t$ as time approaches zero, this is not sufficient to correct errors accumulated earlier in sampling by using an inaccurate score function estimate. For $t<1$, the density induced by solving the reverse SDE from 1 to $t$ using the score sum $\nabla \log \tilde{p}_t$ is not the same as $p^{(\text{full})}_t$ or even $\tilde{p}_t$, since this would still result in the correct density at time 0. This is because $\tilde{p}_t$ is not obtained by the noising SDE whose coefficients are used in the reverse SDE.

The problem of mismatch between score function and induced density is also encountered to some extent in the usual diffusion model sampling procedure, where numerical integration error causes generated samples to not follow the `true' density induced by the reverse SDE. \citet{Song-etal2021a} noticed this and proposed using a step of ULA after each SDE integration step to correct the distribution. They called this predictor-corrector sampling since it uses the MCMC sampler to correct the distribution of the samples proposed by the SDE solver. The annealed sampling procedure described in Section \ref{diffusioncompsection} can be seen as corrector-only sampling in the framework of \cite{Song-etal2021a} and is similar to the method used by \cite{SongErmon2019} for sampling in the discrete-time formulation of the variance exploding SDE.
\FloatBarrier

\end{document}